\documentclass[letterpaper]{article} 
\usepackage{aaai2027}  
\usepackage{times}  
\usepackage{helvet}  
\usepackage{courier}  
\usepackage[hyphens]{url}  
\usepackage{graphicx} 
\usepackage{natbib}  
\usepackage{caption} 
\usepackage{graphicx}
\usepackage{booktabs}

\usepackage{tabularx}
\usepackage{multirow}
\usepackage{makecell}
\usepackage{adjustbox}
\usepackage{float}
\usepackage{colortbl}
\usepackage{booktabs}
\usepackage{multirow}
\usepackage{tabularx}
\usepackage{array}

\newcolumntype{Y}{>{\centering\arraybackslash}X}
\usepackage{amsmath}
\usepackage{paralist}
\usepackage{amsthm}
\usepackage{amssymb}
\nocopyright

\usepackage{algorithm}
\usepackage{algorithmic}

\usepackage{newfloat}
\usepackage{listings}
\DeclareCaptionStyle{ruled}{labelfont=normalfont,labelsep=colon,strut=off} 
\floatstyle{ruled}
\newfloat{listing}{tb}{lst}{}
\floatname{listing}{Listing}
\title{SkillSight: Calibrating Generic Content Bias for Skill Retrieval}
\author{
    Jinying Xiao\textsuperscript{\rm 1},
    Bin Li\textsuperscript{\rm 1},
    Xiaopeng Li\textsuperscript{\rm 1},
    Jianling Li\textsuperscript{\rm 1},
    Jiacheng Jie\textsuperscript{\rm 1},\\
    Xiaodong Liu\textsuperscript{\rm 1},
    Ma Jun\textsuperscript{\rm 1},
    Chao Wang\textsuperscript{\rm 2},
    Nyima Tashi\textsuperscript{\rm 3},
    Jie Yu\textsuperscript{\rm 1}
}

\affiliations{
    \textsuperscript{\rm 1}National University of Defense Technology\\
    \textsuperscript{\rm 2}Qinghai Normal University\\
    \textsuperscript{\rm 3}Xizang University
}

\usepackage{bibentry}

\begin{document}

\maketitle

\begin{abstract}
As large language model agents gain access to increasingly large skill libraries, retrieving the right skill becomes critical to reliable capability selection and execution. 
Existing retrievers often treat skill contents as ordinary documents, overlooking their highly regular structure: shared descriptive patterns recur across many skills while providing little evidence for distinguishing the required capability. 
We show that this shared descriptive background is reflected in dense relevance scores, induces a pronounced energy gap between queries and skill documents, and obscures discriminative signals, especially for structurally similar hard negatives.
Based on this observation, we propose SkillSight, a training-free retrieval framework that calibrates shared background in both semantic and lexical spaces. Semantic Background Calibration estimates a background subspace from generic tokens identified by IDF, reducing similarity induced by shared descriptive patterns, while Lexical Evidence Calibration downweights shared background tokens to recover discriminative token-level evidence.
Experiments on SRA-Bench and SkillBench-Supp demonstrate consistent improvements across retrieval metrics, with SkillSight improving Recall@10 by up to 20.21 percentage points over the original dense retriever. It is up to 1,248 times faster than the Dense + Reranker baseline. In end-to-end evaluation, SkillSight achieves the best overall performance across three agent models and outperforms LLM Selection by up to 4.97 percentage points. These results identify shared descriptive background as a source of ranking interference in skill retrieval and demonstrate that calibrating it enables accurate and efficient skill selection without additional training. Our code can be found at https://github.com/xiaojinying/SkillSight.
\end{abstract}

\begin{figure*}[t]
    \centering
    \includegraphics[width=\textwidth]{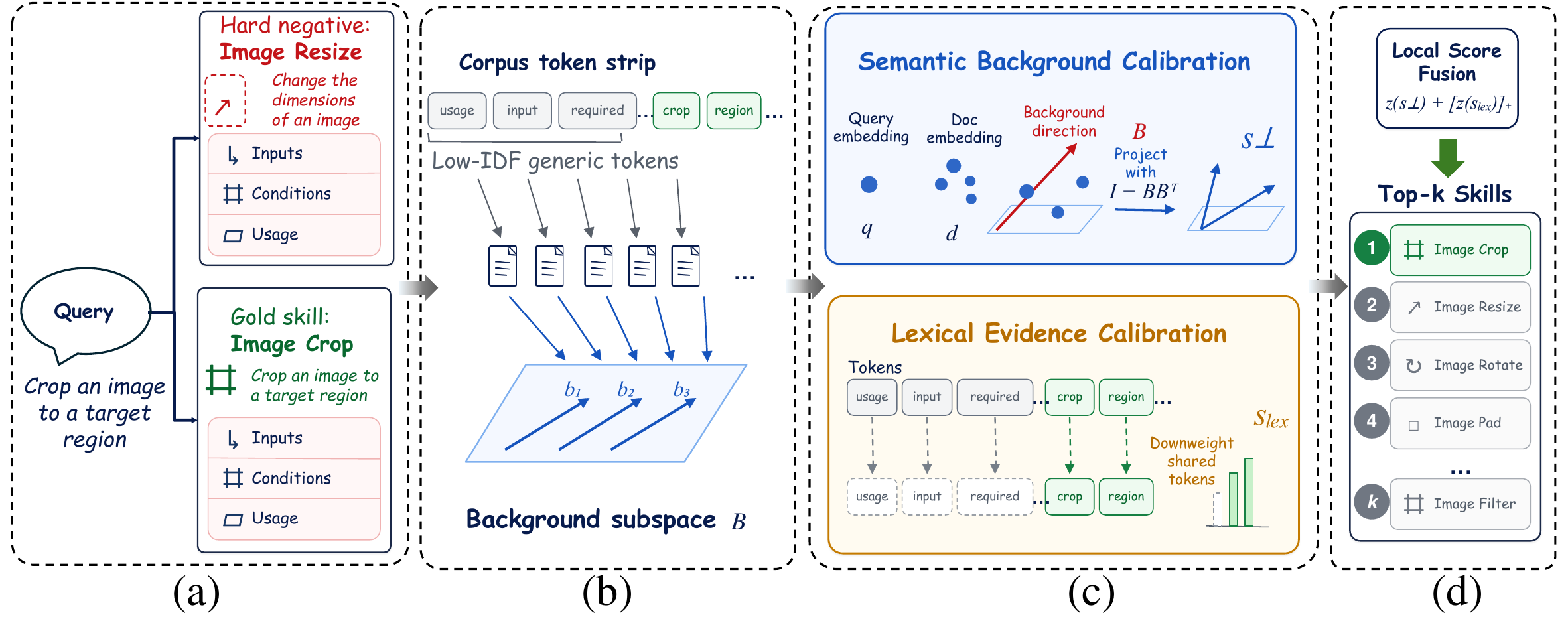} 
    \caption{Overview of SkillSight. (a) The two distinct skills, Image Resize and Image Crop, share substantial descriptive content in their inputs, conditions, and usage instructions. (b) Generic tokens with low IDF values are used to estimate the background subspace $\mathbf{B}$ shared across the skill corpus. (c) Semantic Background Calibration reduces similarity caused by shared background content, while Lexical Evidence Calibration reduces the influence of common tokens and highlights tokens related to the requested capability. (d) The calibrated semantic and lexical scores are normalized and combined to rerank the candidates and return the top $k$ skills.}
    \label{overall_fig}
\end{figure*}

\section{Introduction}
\label{sec:introduction}

As large language model agents gain access to an increasing number of skills~\cite{xu-etal-2026-skill,ding2026skillgen}, accurately identifying the capabilities required for a given task from a large skill repository becomes a critical retrieval problem in agent systems. A skill document typically specifies the capability scope, invocation interface, execution conditions, and usage instructions~\cite{yuan2025easytool}. Providing all documents directly to a large language model incurs substantial context overhead and increases the risk of selecting or invoking irrelevant capabilities~\cite{ICLR2024_28e50ee5}. Skill retrieval addresses this problem by returning a compact set of relevant candidates. The agent then selects and executes the appropriate skill. Retrieval quality therefore directly affects the agent's ability to use external capabilities effectively.

Existing studies improve skill retrieval through query reformulation, specialized retrievers, candidate reranking, and tool-relation modeling~\cite{chen2024re,zheng2024toolrerank,ICLR2024_28e50ee5,qu2024towards}, primarily addressing query--skill semantic mismatch or reassessing candidate relevance with stronger models. However, \textbf{skill retrieval exhibits a distinctive input asymmetry}: queries express instance-specific objectives, targets, and constraints, whereas reusable skill documents follow regular descriptive structures and extensively share content about capability specifications, invocation conditions, and execution procedures (see Figure~\ref{overall_fig}(a)). This structural regularity can induce document-side background bias in dense retrieval, because query--document similarity aggregates task-relevant evidence with shared descriptive content. Consequently, a hard negative may receive a high score by matching generic interfaces or usage instructions despite mismatching the required operation, target, or constraints. Existing representation correction methods remove common embedding directions~\cite{mu2018all,zhou2024ease}, while lexical augmentation and interaction-based reranking introduce complementary matching signals~\cite{bruch2023analysis,zhuang2024promptreps}. However, neither explicitly identifies shared skill contents as a semantic source of ranking bias nor quantifies their contribution to relevance scores.

We analyze this problem from three perspectives: ranking behavior, token statistics, and representation geometry. First, compared with conventional text retrieval datasets~\cite{thakur2beir}, skill retrieval datasets require a greater retrieval depth to achieve the same coverage of gold skills, and a larger proportion of candidate documents are ranked ahead of the gold skill. Second, generic tokens occur across many skill documents, but only a small fraction of the documents containing them correspond to the gold skill for a given query. We further use generic tokens to estimate the background subspace of a skill corpus and observe that skill documents consistently exhibit higher background energy than queries in this subspace. This query--document energy gap is not evident in conventional text retrieval datasets. An orthogonal decomposition of the dense score further shows that background alignment contributes additional scores to some candidates, thereby reducing the influence of task-specific evidence on ranking.

Based on these findings, we propose SkillSight, a training-free relevance calibration framework for skill retrieval. Semantic Background Calibration (SBC) estimates a corpus-level background subspace from generic tokens and removes its contribution to dense similarity, while Lexical Evidence Calibration (LEC) downweights corpus-common terms to emphasize fine-grained capability-matching evidence. The calibrated semantic and lexical scores are fused within a local candidate set without additional training or model inference.

Across SRA-Bench~\cite{su2026skill} and SkillBench-Supp~\cite{zheng2026skillrouter}, SkillSight improves Recall@10 by up to 20.21 points and achieves the best non-oracle performance across three agent models, while remaining up to 1,248$\times$ faster than neural reranking.

Our contributions are as follows:
\begin{itemize}
    \item We identify document-side background bias in skill retrieval, showing that shared descriptive patterns systematically enter dense relevance scores and induce a query-document energy gap that weakens matching.

    \item We introduce SkillSight, a training-free framework that calibrates corpus-shared background across semantic and lexical relevance signals to emphasize capability specific evidence, requiring no auxiliary model inference.

    \item We demonstrate that SkillSight consistently improves retrieval across datasets and embedding models, translates these gains into better downstream agent execution, and remains substantially faster than neural reranking.
\end{itemize}

\section{Related Work}
\label{sec:related_work}

\subsection{Agent Tool and Skill Retrieval}

Large tool libraries have motivated retrieval-based selection.
ToolLLM introduces a dedicated API retriever~\cite{ICLR2024_28e50ee5},
while subsequent work improves tool retrieval through query
reformulation, reranking, and relation
modeling~\cite{chen2024re,zheng2024toolrerank,qu2024towards}.
Recent studies extend this setting to skill retrieval and agent
execution~\cite{shi2025retrieval,li2026skillsbench,su2026skill}.
SkillRouter trains a skill-specific bi-encoder and reranker,
whereas SkillRet develops a large-scale benchmark and studies
task-specific retriever adaptation~\cite{zheng2026skillrouter,
cho2026skillret}. These methods primarily improve skill matching
through specialized supervision or model training. In contrast,
SkillSight identifies corpus-shared descriptive patterns as a
source of dense-score bias and calibrates existing retrievers
without additional training or neural inference.

Graph-of-Skills and Group-of-Skills organize or compose related skills before retrieval~\cite{liu2026graph,zeng2026group}. SkillSight instead studies background-induced bias in flat, individual-skill ranking, making these structured retrieval settings not directly comparable.

\subsection{Retrieval Representations and Relevance Calibration}

Dense retrieval estimates relevance from the overall similarity
between query and document representations~\cite{karpukhin2020dense},
which may underemphasize fine-grained evidence such as operations,
entities, and constraints. Hybrid methods supplement dense
representations with lexical signals~\cite{zhuang2024promptreps,
bruch2023analysis}, but conventional fusion does not distinguish
capability-specific evidence from recurrent template language.
SkillSight explicitly calibrates corpus-shared background in the
semantic channel and introduces discriminative lexical evidence
for capability matching.

Representation post-processing methods reduce non-discriminative
components using word frequency or global embedding
geometry~\cite{arora2017simple,mu2018all,zhou2024ease}. Their
objective is typically to improve general representation quality
or isotropy. SkillSight instead grounds the background subspace
in generic-token--document associations within the skill corpus
and targets its asymmetric contribution to query--document
scoring. This task-specific attribution distinguishes SkillSight
from general geometric correction.

\section{Analysis}

\subsection{Shared Descriptive Background}

A dense retriever uses the inner product between the query and document embeddings as the relevance score:
\begin{equation}
s(q,d)=\mathbf{q}^{\top}\mathbf{d}.
\label{eq}
\end{equation}

This score aggregates all alignment signals in the embedding space. In skill retrieval, these signals can be broadly divided into two categories. The first captures task-specific evidence required by the query, whereas the second reflects the explanatory background shared across skill documents. The former determines whether a skill actually satisfies the current request, while the latter mainly indicates whether a document follows a typical skill descriptive pattern.

\begin{figure}[t]
\centering
\includegraphics[width=1\linewidth]{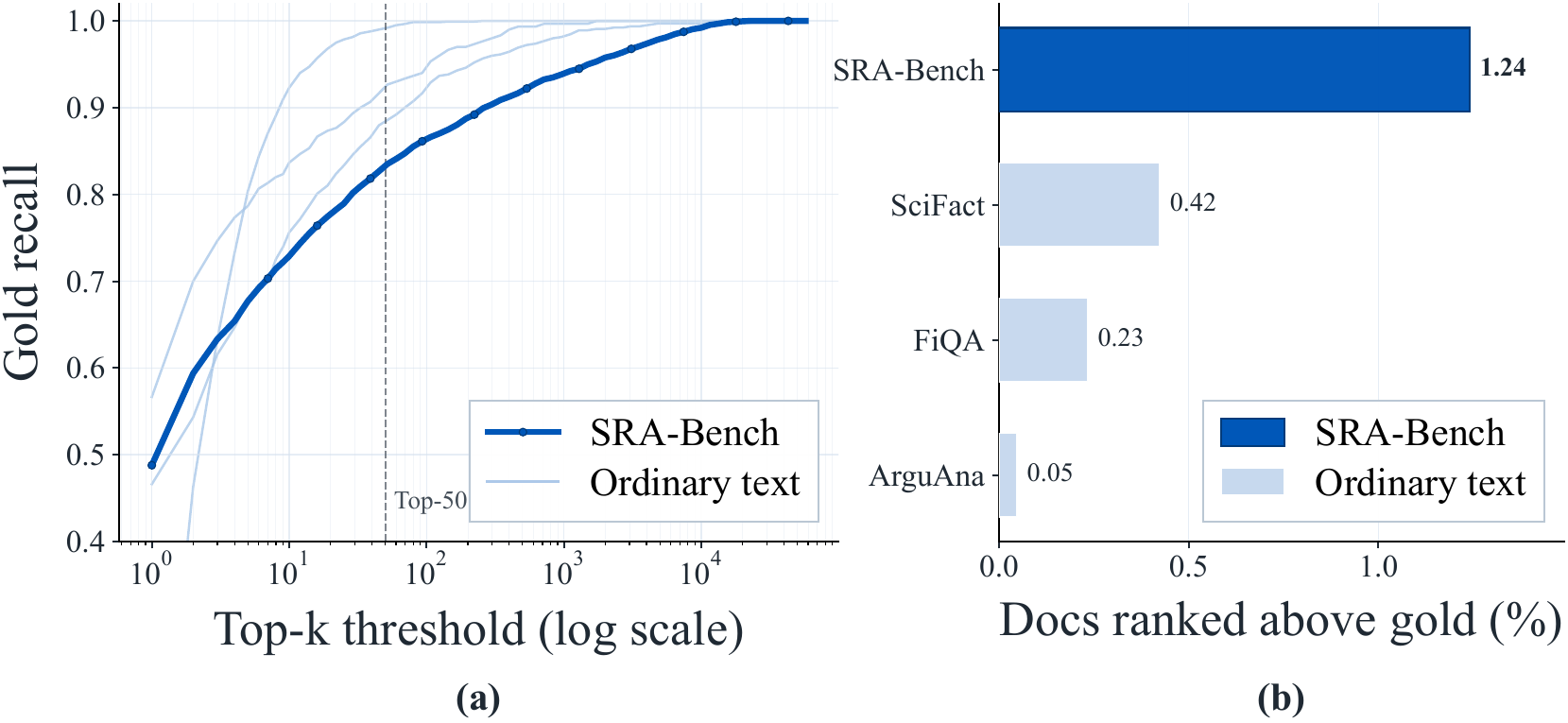}
\caption{
Retrieval ranking behavior for SRA-Bench and ordinary-text benchmarks using Qwen3-Embedding-0.6B with cosine similarity. 
(a) Gold recall over Top-k retrieval depth on a log scale, where gold recall is the fraction of queries whose best gold document appears within the top k results. 
(b) Average percentage of corpus documents ranked above the best gold document, computed as mean $((rank-1) / corpus \ size) \times 100$.
}
\label{fig:1}
\end{figure}

This mixture of signals affects the ranking of hard negatives. Figure~\ref{fig:1} shows that the recall curve for gold skills on SRA-Bench consistently lags behind that observed in conventional text retrieval~\cite{thakur2beir}, with a larger average proportion of candidate documents ranked ahead of the gold document. Specifically, a hard negative may resemble the gold skill in writing style, interface content, and semantic structure, thereby receiving a high dense score, even though it mismatches the query in key operations, target entities, or domain conditions. Such background alignment weakens the ranking contribution of fine-grained capability-matching signals and introduces a systematic bias into skill retrieval.
\begin{figure}[t]
\centering
\includegraphics[width=1\linewidth]{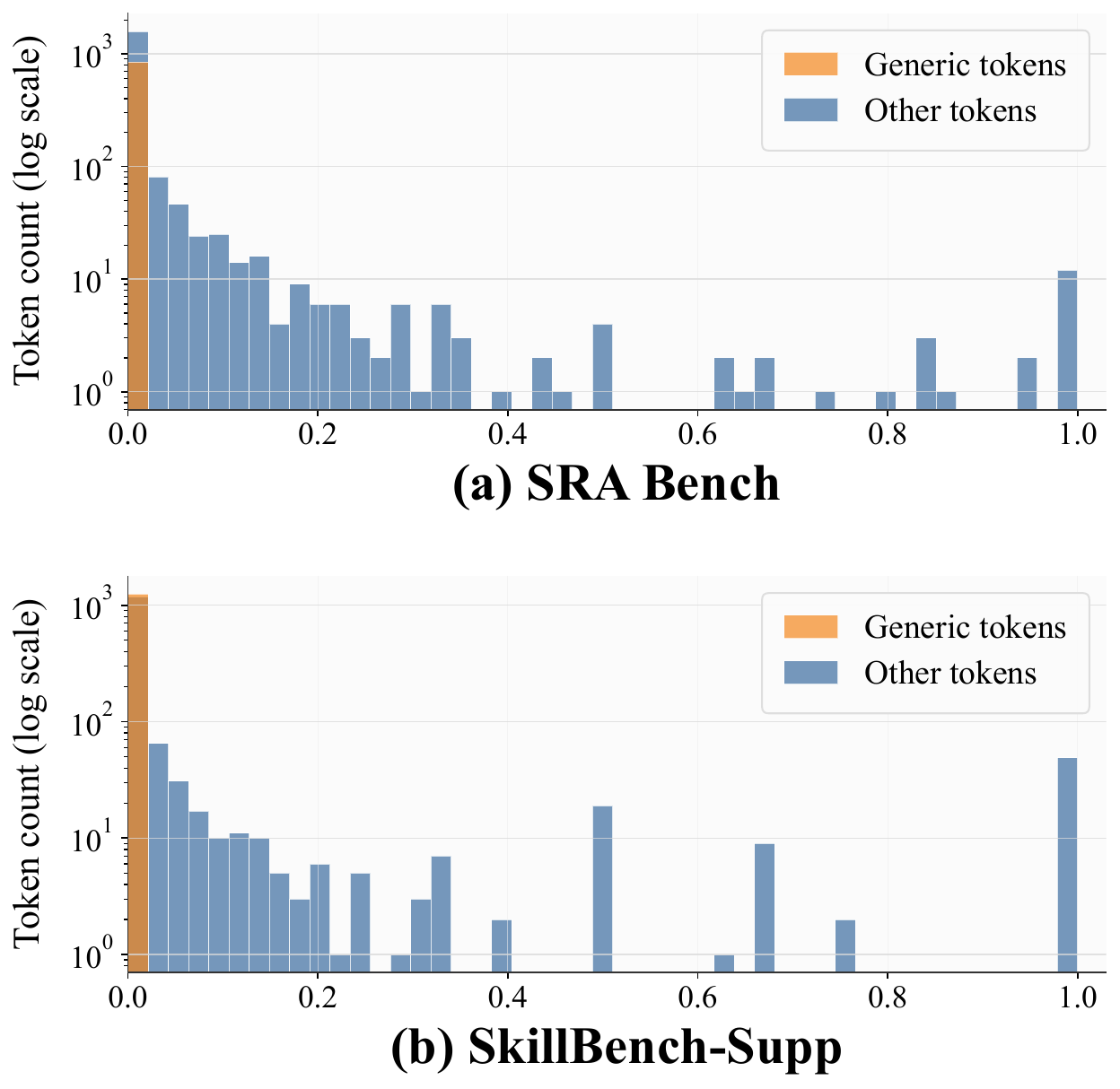}
\caption{
Gold-skill rate distributions of different token groups on SRA-Bench and SkillBench-Supp. The horizontal axis represents the proportion of skills containing a given token that serve as the gold skill, while the vertical axis shows the corresponding number of tokens on a log scale.
}
\label{fig:2}
\end{figure}

\subsection{Query--Document Background Energy Gap}
\label{sec:energy_gap}

We characterize the shared background of a skill corpus through token commonality. Let $\mathcal{D}=\{d_i\}_{i=1}^{N}$ denote the skill documents with normalized embeddings $\{\mathbf{d}_i\}_{i=1}^{N}$. For each corpus token $t$, we define
\begin{equation}
\operatorname{idf}(t)
=
\log\frac{N+1}{\operatorname{df}(t)+1}+1,
\label{eq:idf}
\end{equation}
where $\operatorname{df}(t)$ is the number of documents containing $t$. We apply Otsu's method~\cite{4310076} to the IDF distribution and denote the resulting low-IDF token set by $\mathcal{T}_g$. Figure~\ref{fig:2} shows that the gold-skill rates of these generic tokens concentrate near zero, whereas other tokens retain greater mass in higher-rate regions. This indicates that corpus frequent tokens primarily encode shared contents and provide limited discrimination among skills.

We map this token-level commonality into the embedding space. For each $t\in\mathcal{T}_g$, let $\mathcal{I}_t=\{i\mid t\in d_i\}$ and compute its associated mean document embedding:
\begin{equation}
\mathbf{g}_t
=
\frac{1}{|\mathcal{I}_t|}
\sum_{i\in\mathcal{I}_t}\mathbf{d}_i.
\label{eq:token_mean}
\end{equation}

We collect these token-induced directions together with the corpus mean $\bar{\mathbf{d}}=\frac{1}{N}\sum_{i=1}^{N}\mathbf{d}_i$ and perform singular value decomposition:
\begin{equation}
G=
\left[
\mathbf{g}_{t_1},\ldots,\mathbf{g}_{t_m},
\bar{\mathbf{d}}
\right]
=
U\Sigma V^\top,
\quad
B=[\mathbf{u}_1,\ldots,\mathbf{u}_r],
\label{eq:background_subspace}
\end{equation}
where $\{t_1,\ldots,t_m\}\subseteq\mathcal{T}_g$, and $r$ denotes the
selected rank of the background subspace. The orthonormal basis $B$ spans a low-dimensional subspace that summarizes the background directions shared across skill documents.

For any normalized embedding $\mathbf{v}$, we define its background energy as
\begin{equation}
e_B(\mathbf{v})
=
\left\|B^\top\mathbf{v}\right\|_2^2.
\label{eq:background_energy}
\end{equation}

Given normalized query embeddings $\{\mathbf{q}_j\}_{j=1}^{M}$, the average document- and query-side energies are
\begin{equation}
E_D
=
\frac{1}{N}\sum_{i=1}^{N}e_B(\mathbf{d}_i),
\qquad
E_Q
=
\frac{1}{M}\sum_{j=1}^{M}e_B(\mathbf{q}_j).
\label{eq:mean_background_energy}
\end{equation}

\begin{figure}[t]
\centering
\includegraphics[width=1\linewidth]{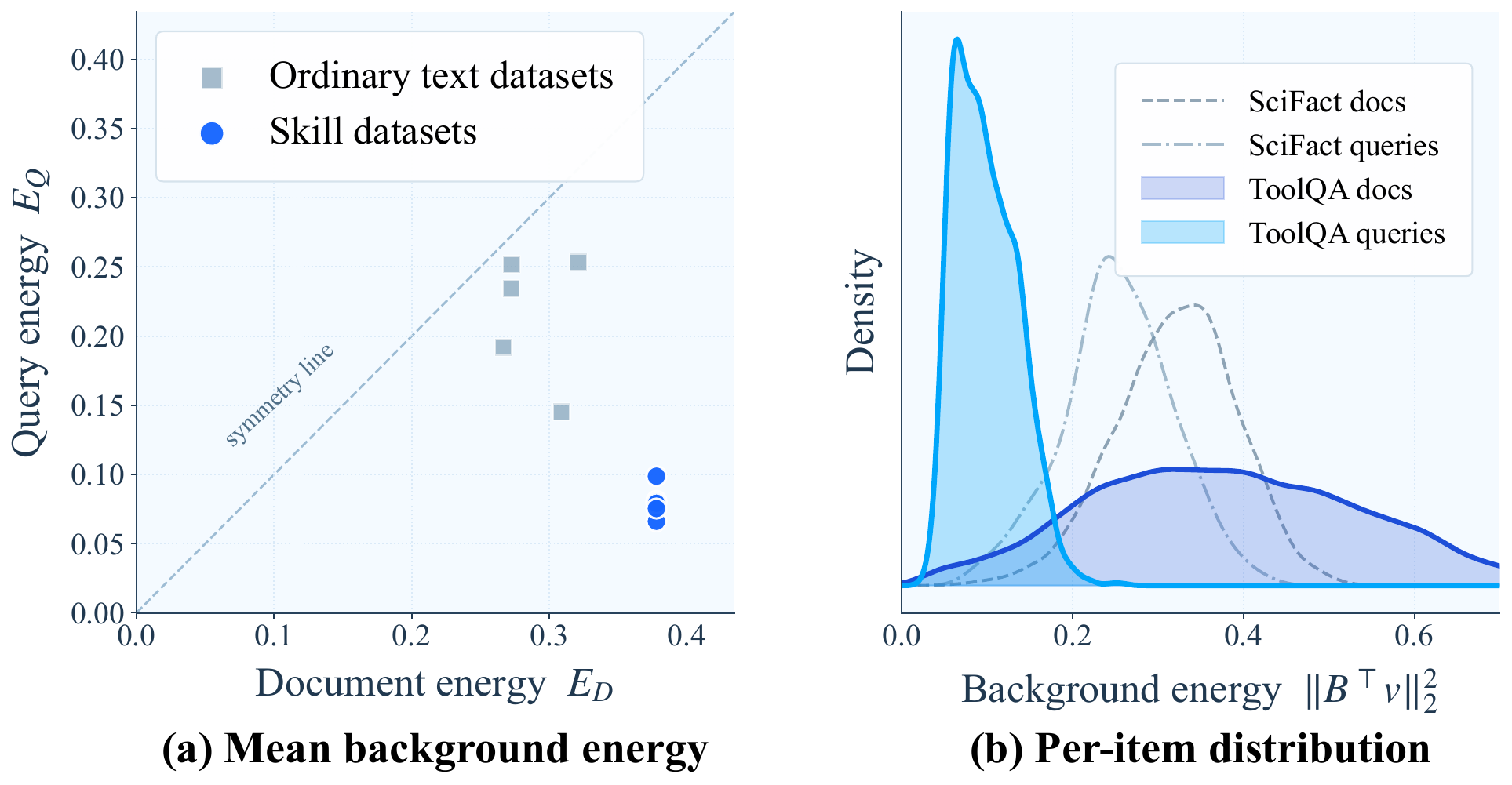}
\caption{
Background energy statistics for queries and documents.
(a) Average background energy across datasets. The horizontal axis denotes document energy, and the vertical axis denotes query energy. The dashed line indicates equal query and document energy, while different markers distinguish skill datasets from ordinary text corpora.
(b) Distributions of background energy for individual queries and documents. The horizontal axis denotes background energy, and the vertical axis denotes density. Different curves represent the queries and documents in SciFact and ToolQA.
}
\label{fig:3}
\end{figure}

As shown in Figure~\ref{fig:3}, conventional text retrieval datasets lie close to the symmetry line $E_D=E_Q$, whereas skill retrieval datasets consistently exhibit $E_D>E_Q$. Their document-level energy distributions also shift toward higher-energy regions, showing that stronger coupling to the background subspace is a systematic property of skill documents.

This energy gap directly affects dense relevance scoring. Let $P_B=BB^\top$ be the projection matrix onto the background subspace. Since $P_B$ and $I-P_B$ project onto orthogonal subspaces, the dense score decomposes as
\begin{equation}
\mathbf{q}^\top\mathbf{d}
=
\underbrace{(P_B\mathbf{q})^\top(P_B\mathbf{d})}_{\text{background alignment}}
+
\underbrace{\bigl((I-P_B)\mathbf{q}\bigr)^\top
\bigl((I-P_B)\mathbf{d}\bigr)}_{\text{residual alignment}}.
\label{eq:score_decomposition}
\end{equation}

When documents exhibit elevated background energy, the first term can provide additional scores to candidates whose shared contents align with the query. Because this contribution is only weakly related to the required capability, it can obscure task-specific evidence and bias the resulting ranking.

\section{SkillSight}
\label{sec:SkillSight}
Based on this observation, SkillSight calibrates shared corpus information at both the semantic representation and lexical matching levels. The semantic channel removes alignment within the background subspace, while the lexical channel reduces the matching contribution of frequently shared tokens. SkillSight then combines these two complementary sources of relevance evidence within the candidate set.

\subsection{Semantic Background Calibration}
\label{sec:background_calibrated_dense}

Given an orthonormal basis $B$ of the background subspace, its projection matrix is $P_B=BB^\top$. SkillSight projects both the query and document representations onto the orthogonal complement of the background subspace:
\begin{equation}
\mathbf{q}_{\perp}
=
(I-P_B)\mathbf{q},
\qquad
\mathbf{d}_{\perp}
=
(I-P_B)\mathbf{d}.
\label{eq:residual_projection}
\end{equation}

The background-calibrated dense relevance score is defined as
\begin{equation}
s_{\perp}(q,d)=\mathbf{q}_{\perp}^{\top}\mathbf{d}_{\perp}.
\label{eq:projected_dense_score}
\end{equation}

According to Eq.~\ref{eq:score_decomposition}, $s_{\perp}(q,d)$ corresponds to the semantic alignment outside the background subspace and therefore excludes the direct contribution of background alignment. We do not renormalize the representations after projection, allowing their norms to preserve the residual semantic energy of the original embeddings. For a document with high background energy, the remaining component available for relevance matching becomes correspondingly smaller, which reduces the influence of shared descriptive patterns on the ranking.

After background calibration, the dense score provides a more task-specific estimate of semantic relevance. Figure~\ref{fig:2} further shows that other tokens retain substantial mass in regions with high gold-skill fractions, suggesting that lexical signals contain additional discriminative information. SkillSight therefore incorporates lexical evidence to improve the distinction between specific operations, entities, and constraints.

\subsection{Lexical Evidence Calibration}
\label{sec:query_specific_lexical}

We first remove the generic tokens from the query token set:
\begin{equation}
\mathcal{T}_q^{\mathrm{sp}}
=
\mathcal{T}_q\setminus\mathcal{T}_g.
\label{eq:specific_query_tokens}
\end{equation}

As shown in Figure~\ref{fig:2}, the remaining tokens still vary substantially in discriminative power. We quantify their residual commonality by document prevalence,
\begin{equation}
p(t)=\frac{\mathrm{df}(t)}{N},
\label{eq:token_prevalence}
\end{equation}
and assign each $t\in\mathcal{T}_q^{\mathrm{sp}}$ the calibrated weight
\begin{equation}
w_{\beta}(t)
=
\mathrm{idf}(t)\bigl(1-p(t)\bigr)^{\beta},
\qquad \beta\geq 0.
\label{eq:qcle_weight}
\end{equation}

The factor $(1-p(t))^\beta$ further suppresses tokens shared by many documents, with $\beta$ controlling the calibration strength. When $\beta=0$, the weight reduces to standard IDF.

For a candidate document $d$, we define the lexical relevance score as the weighted coverage of non-generic query tokens:
\begin{equation}
s_{\mathrm{lex}}(q,d)
=
\frac{
\displaystyle
\sum_{t\in\mathcal{T}_q^{\mathrm{sp}}}
w_{\beta}(t)\mathbb{I}[t\in d]
}{
\displaystyle
\sum_{t\in\mathcal{T}_q^{\mathrm{sp}}}
w_{\beta}(t)
}.
\label{eq:qcle_score}
\end{equation}

This score emphasizes discriminative exact matches and complements the capability-level semantic evidence captured by the dense channel.

Because $s_{\perp}$ and $s_{\mathrm{lex}}$ have different scales, we normalize each channel $c\in\{\perp,\mathrm{lex}\}$ within the candidate set $\mathcal{C}_q$:
\begin{equation}
\widetilde{s}_{c}(q,d)
=
\frac{s_c(q,d)-\mu_c(q)}
{\sigma_c(q)+\epsilon},
\label{eq:local_score_normalization}
\end{equation}
where $\mu_c(q)$ and $\sigma_c(q)$ are the corresponding mean and standard deviation over $\mathcal{C}_q$. The final relevance score is
\begin{equation}
s_{\mathrm{SkillSight}}(q,d)
=
\widetilde{s}_{\perp}(q,d)
+
\left[\widetilde{s}_{\mathrm{lex}}(q,d)\right]_{+},
\label{eq:SkillSight_final_score}
\end{equation}
where $[x]_{+}=\max(x,0)$. The background-calibrated dense score provides the primary ranking signal, while above-average lexical matches receive a positive correction.

\begin{table*}[bth]
\centering
\begin{tabularx}{\textwidth}{@{}l*{8}{Y}@{}}
\toprule
\multirow{2}{*}{Method}
& \multicolumn{4}{c}{SRA-Bench}
& \multicolumn{4}{c}{SkillBench-Supp} \\
\cmidrule(lr){2-5} \cmidrule(l){6-9}
& Hit@5 & Recall@10 & MRR@10 & Latency
& Hit@5 & Recall@10 & MRR@10 & Latency\\
\midrule
BM25                  & 63.37 & 63.51 & 51.54 & 3.41 & 65.75 & 53.15 & 57.16 & 24.20 \\
TF-IDF Cosine         & 56.30 & 58.04 & 42.56 & 2.32 & 53.42 & 42.11 & 46.26 & 16.42 \\
ColBERTv2             & 41.88 & 42.36 & 31.36 & 12.42 & 56.16 & 38.94 & 38.69 & 31.54 \\
SPLADE                & 55.25 & 54.91 & 43.60 & 40.55 & 28.77 & 14.40 & 22.88 & 42.01 \\
ABTT                  & 67.70 & 66.01 & 57.08 & 0.12 & 75.34 & 59.77 & 66.53 & 3.18 \\
SIF                   & 35.46 & 36.07 & 27.28 & 0.11 & 39.73 & 29.02 & 33.44 & 0.75 \\
Dense   & 67.70 & 66.02 & 57.09 & 0.13 & 75.34 & 56.56 & 66.67 & 0.57 \\
BGE-M3 Hybrid         & 65.67 & 65.46 & 55.29 & 19.07 & 72.60 & 58.60 & 62.20 & 45.82 \\
Dense--BM25 RRF       & 75.67 & 77.13 & 62.10 & 167.74 & 73.97 & 56.72 & 64.36 & 219.26 \\
Dense + Reranker  & 80.87 & 77.18 & 72.21 & 1460.35 & \textbf{76.71} & 56.83 & 66.71 & 1740.79 \\
\midrule
\textbf{SkillSight}                  & \textbf{86.04} & \textbf{86.23} & \textbf{74.02} & 1.17
                      & \textbf{76.71} & \textbf{64.24} & \textbf{67.28} & 2.57 \\
\bottomrule
\end{tabularx}
\caption{Main retrieval results on SRA-Bench and SkillBench-Supp. Dense, ABTT, SIF, and SkillSight use Qwen3-Embedding-0.6B, while the reranker uses Qwen3-Reranker-0.6B. Latency is measured in milliseconds (ms). The best result for each retrieval metric is highlighted in \textbf{bold}.} 
\label{tab:1}
\end{table*}

\section{Experiments}
\subsection{Experimental Setup}
\label{sec:experimental_setup}

\paragraph{Datasets}
We conduct retrieval experiments on SRA-Bench~\cite{su2026skill}
and SkillBench-Supp~\cite{zheng2026skillrouter}, both of which
provide queries, candidate skill documents, and corresponding
gold skills. SRA-Bench contains 5,400 test instances and
approximately 26k candidate skills, whereas SkillBench-Supp
contains approximately 77k candidate skills, covering retrieval
settings at different corpus scales. We additionally report
comparative retrieval results on SkillRet~\cite{cho2026skillret} in Appendix~C of the
supplementary material. End-to-end agent evaluation is conducted
on SRA-Bench, whose unified execution protocol enables controlled
comparison across retrieval methods.

\paragraph{Baselines}
We compare SkillSight with representative retrieval methods covering complementary relevance signals. BM25~\cite{robertson2009probabilistic} and SPLADE~\cite{formal2021splade} represent sparse retrieval, Dense~\cite{zhang2025qwen3} represents dense retrieval, and ColBERTv2~\cite{santhanam2022colbertv2} models fine-grained token interactions. We further include ABTT~\cite{mu2018all} for representation-space correction, BGE-M3 Hybrid~\cite{chen-etal-2024-m3} for dense--sparse fusion, and Dense + Reranker for two-stage ranking. For end-to-end evaluation, we compare against LLM Selection and Progressive Disclosure, two representative skill-incorporation strategies evaluated in SRA-Bench~\cite{su2026skill}. LLM Selection uses the agent model to select a skill from the retrieved candidates before execution, whereas Progressive Disclosure allows the agent to load full skill content on demand.

\paragraph{Evaluation Metrics}
For offline retrieval, we report Hit@$k$, Recall@$k$, and MRR@$k$, with formal definitions provided in Appendix B. For end-to-end evaluation, we measure agent execution performance according to the task-specific evaluation protocol of each benchmark. All methods use the same agent model and execution environment, with only the skill retrieval or selection module varied.

\paragraph{Implementation Details}
For each query, SBC constructs an intermediate candidate pool $\mathcal{C}_q$ of size $K_c=300$, within which semantic and lexical evidence are fused for final ranking. We set $\beta=1$ and determine $r$ automatically from the spectral effective rank of the background matrix without evaluation labels. For end-to-end evaluation, the top three retrieved skill documents are inserted into the context of Llama-3.1-8B-Instruct~\cite{grattafiori2024llama}, GPT-5.4-mini, or Qwen3-4B-Instruct~\cite{yang2025qwen3}. Appendices~A and~D provide the algorithmic details and hyperparameter analyses, respectively. The code included in the supplementary material provides a complete workflow for reproducing all experimental results.

\begin{table*}[bth]
\centering
\label{tab:main_results}
\resizebox{\textwidth}{!}{
\begin{tabular}{llccccccc|cc}
\toprule
\textbf{Model} & \textbf{Method} 
& \textbf{TheoremQA} 
& \textbf{LogicBench} 
& \textbf{ToolQA} 
& \textbf{CHAMP} 
& \textbf{MedCalc} 
& \textbf{BigCodeBench} 
& \textbf{Overall}
& \textbf{Latency (s)}
& \textbf{Token} \\
\midrule

\multirow{5}{*}{Llama-3.1-8B}
& LLM Direct              & 31.06 & 54.47 & 3.29 & 22.42 & 27.55 & 30.09 & 25.72 & 10.42 & 4,496 \\
& Oracle Skill           & 46.18 & 68.95 & 16.85 & 39.46 & 60.82 & 36.14 & 42.20 & 13.05 & 6,617 \\
\cmidrule(lr){2-11}
& LLM Selection          & 35.34 & 54.87 & 4.69 & 27.80 & 56.36 & \textbf{34.82} & 33.83 & 21.65 & 8,525 \\
& Progressive Disclosure & 38.69 & 48.42 & 0.35 & 26.46 & \textbf{58.82} & 31.14 & 31.91 & 74.18 & 10,942 \\
& \textbf{SkillSight}                   & \textbf{42.70} & \textbf{61.05} & \textbf{14.90} & \textbf{30.94} & 58.64 & 33.77 & \textbf{38.80} & 19.91 & 7,703 \\
\midrule

\multirow{5}{*}{GPT-5.4-mini}
& LLM Direct              & 78.31 & 76.84 & 41.05 & 77.58 & 77.36 & 49.12 & 61.85 & 5.51 & 4,581 \\
& Oracle Skill           & 81.93 & 92.11 & 55.24 & 81.17 & 90.64 & 61.40 & 73.70 & 6.03 & 6,521 \\
\cmidrule(lr){2-11}
& LLM Selection          & 79.92 & 80.13 & \textbf{48.25} & 79.37 & 86.82 & 55.70 & 67.83 & 17.02 & 10,591 \\
& Progressive Disclosure & 79.92 & 73.55 & 44.20 & 79.82 & 89.45 & 53.42 & 65.91 & 7.87 & 10,258 \\
& \textbf{SkillSight}                   & \textbf{81.39} & \textbf{84.61} & 45.66 & \textbf{82.06} & \textbf{89.82} & \textbf{60.26} & \textbf{69.67} & 6.12 & 9,205 \\
\midrule

\multirow{5}{*}{Qwen3-4B}
& LLM Direct              & 51.27 & 76.32 & 27.41 & 68.61 & 36.18 & 42.41 & 44.25 & 13.58 & 3,279 \\
& Oracle Skill           & 67.47 & 88.68 & 47.62 & 71.75 & 82.91 & 49.30 & 64.69 & 16.86 & 5,985 \\
\cmidrule(lr){2-11}
& LLM Selection          & 64.66 & 77.63 & 35.59 & 69.96 & 77.18 & 42.37 & 56.85 & 26.85 & 8,829 \\
& Progressive Disclosure & 66.67 & 79.34 & 29.23 & 70.85 & 67.73 & 42.32 & 53.79 & 30.85 & 9,446 \\
& \textbf{SkillSight}                 & \textbf{68.40} & \textbf{80.53} & \textbf{41.05} & \textbf{73.99} & \textbf{80.91} & \textbf{47.02} & \textbf{61.13} & 20.51 & 8,234 \\

\bottomrule
\end{tabular}
}
\caption{End-to-end performance on SRA-Bench across six tasks and three agent models. Oracle Skill directly provides the annotated gold skills as an reference. From the top-50 BM25 candidates, LLM Selection loads one skill before execution, whereas Progressive Disclosure loads skills on demand during inference. Overall is averaged over all instances, and Token denotes the average number of input tokens per instance. The best non-oracle results are bolded.}
\label{tab:2}
\end{table*}

\subsection{Main Results}
\paragraph{Offline Retrieval}
As shown in Table~\ref{tab:1}, SkillSight consistently improves retrieval performance on both datasets, with the largest gains observed in $\mathrm{Recall}@10$. Compared with the original Dense retriever, SkillSight increases $\mathrm{Recall}@10$ from 66.02 to 86.23 on SRA-Bench and from 56.56 to 64.24 on SkillBench-Supp. SkillSight improves all metrics on SRA-Bench, whereas its gains on SkillBench-Supp are concentrated primarily in recall. This difference suggests that its main effect is to increase the coverage of gold skills among the top-ranked candidates, while improvements at earlier ranks depend more strongly on dataset characteristics. Further retrieval results across different cutoffs and datasets are provided in Appendix~C of the supplementary material.
\paragraph{End-to-end Agent Execution}
As shown in Table~\ref{tab:2}, SkillSight achieves the best non-oracle Overall score across all three models, outperforming LLM Selection by 4.97, 1.84, and 4.28 percentage points, respectively. The improvement is particularly pronounced on ToolQA with weaker models, reaching 10.21 percentage points for Llama-3.1-8B and 5.46 percentage points for Qwen3-4B. This result indicates that when candidate skills share similar functional contents but differ in specific operations and invocation constraints, SkillSight identifies the target skill more accurately by suppressing bias from shared contents and strengthening fine-grained matching signals.

\subsection{Efficiency Analysis}
SkillSight adds limited overhead to standard dense retrieval.
The background subspace $B$ and projected document representations
$d_{\perp}$ are precomputed offline. At inference time, query
projection costs $\mathcal{O}(dr)$, while lexical matching over
$C_q$ costs $\mathcal{O}(|C_q||T_q^{\mathrm{sp}}|)$. The
corpus-wide semantic search retains the standard dense retrieval
cost of $\mathcal{O}(Nd)$. Therefore, SkillSight introduces only
$\mathcal{O}\!\left(dr+|C_q||T_q^{\mathrm{sp}}|\right)$
additional online complexity, without extra neural inference beyond
the base encoder or trainable parameters.

As shown in Table~\ref{tab:1}, SkillSight achieves retrieval latencies of only 1.17 and 2.57 on the two datasets, respectively. It is 85--143$\times$ faster than Dense--BM25 RRF and 677--1,248$\times$ faster than Dense + Reranker, while also achieving better retrieval performance. In the end-to-end setting, Table~\ref{tab:2} shows that SkillSight reduces latency by 8.0\%--64.0\% and token consumption by 6.7\%--13.1\% compared with LLM Selection.

\subsection{Comparison with Specialized Skill Retrievers}
\label{sec:specialized_retrievers}

\paragraph{Full-Pipeline Comparison.}
As shown in Table~\ref{tab:skillrouter_comparison}, SkillSight
outperforms SkillRouter~\cite{zheng2026skillrouter} by 15.84 Recall@10 points and is
$1{,}397\times$ faster on SRA-Bench. On SkillBench-Supp,
SkillRouter leads by 2.13 points, whereas SkillSight is
$716\times$ faster, demonstrating a favorable
accuracy--efficiency trade-off. Since the pretrained SkillRet~\cite{cho2026skillret}
reranker checkpoint was unavailable, we evaluate only its
released embedding model.

\paragraph{Generalization across Embedding Models.}
Table~\ref{tab:embedding_models} shows that SkillSight
consistently improves Recall@10 across Qwen, SkillRouter,
and SkillRet embeddings, with gains of 8.21--21.80 points on
SRA-Bench and 2.62--7.68 points on SkillBench-Supp. This
demonstrates that SkillSight complements retrievers of
varying strengths and is not tied to a particular embedding
model.

\begin{table}[thb]
\centering
\small
\setlength{\tabcolsep}{4.5pt}
\begin{tabular}{lcc|cc}
\toprule
& \multicolumn{2}{c|}{SRA-Bench}
& \multicolumn{2}{c}{SkillBench-Supp} \\
\cmidrule(lr){2-3}\cmidrule(lr){4-5}
Method
& R@10 & Latency
& R@10 & Latency \\
\midrule
SkillRouter
& 70.39 & 1633.97
& 66.37 & 1839.97 \\
SkillSight
& 86.23 & 1.17
& 64.24 & 2.57 \\
\bottomrule
\end{tabular}
\caption{Recall@10 and retrieval latency (ms) of
SkillSight and the complete SkillRouter retrieval pipeline.
SkillRouter uses its specialized embedding and reranking
models, whereas SkillSight requires no neural reranking.}
\label{tab:skillrouter_comparison}
\end{table}

\begin{table}[thb]
\centering
\small
\setlength{\tabcolsep}{4.5pt}
\begin{tabular}{lcccc}
\toprule
& \multicolumn{2}{c}{SRA-Bench}
& \multicolumn{2}{c}{SkillBench-Supp} \\
\cmidrule(lr){2-3}\cmidrule(lr){4-5}
Model
& Dense & SkillSight
& Dense & SkillSight \\
\midrule
Qwen3 & 66.02 & \textbf{86.23} & 56.56 & \textbf{64.24} \\
SkillRouter & 58.01 & \textbf{79.81} & 62.39 &\textbf{68.85}  \\
SkillRet & 78.21 & \textbf{86.42} & 61.78 & \textbf{64.40} \\
\bottomrule
\end{tabular}
\caption{Recall@10 of Dense and SkillSight with Qwen3-Embedding-0.6B, SkillRouter-Embedding-0.6B, and SkillRet-Embedding-0.6B on SRA-Bench and SkillBench-Supp.}
\label{tab:embedding_models}
\end{table}

\begin{table}[hbt]
    \centering
    
    \begin{tabular}{lccc}
        \toprule
        \textbf{Method}
        & \textbf{Hit@5} 
        & \textbf{R@10} 
        & \textbf{MRR@10}  \\
        \midrule
        
        \multicolumn{4}{l}{\textit{Dense retrieval}} \\
        Dense
        & 67.70
        & 66.02
        & 57.09 \\
        
        Dense + LEC
        & 81.50
        & 79.76
        & 69.53 \\
        
        \midrule
        \multicolumn{4}{l}{\textit{Sparse retrieval}} \\
        BM25
        & 63.37
        & 63.51
        & 51.54 \\
        
        BM25 + SBC
        & 72.26
        & 75.41
        & 58.62 \\
        
        \midrule
        \multicolumn{4}{l}{\textit{Ablation study}} \\
        w/o LEC (SBC only)
        & 78.61
        & 78.23
        &65.77 \\
        
        w/o SBC (LEC only)
        & 69.43
        & 71.46
        & 54.28 \\
        
        \midrule
        \textbf{SkillSight}
        & \textbf{86.04}
        & \textbf{86.23}
        & \textbf{74.02} \\
        
        \bottomrule
    \end{tabular}%
    \caption{Component-level ablation results on SRA-Bench.
Dense + LEC and BM25 + SBC evaluate the incremental
contribution of lexical and semantic calibration under dense
and sparse retrieval paradigms, respectively. The remaining
variants remove the corresponding component from the complete
SkillSight framework. }
\label{tab:ablation}
\end{table}

\begin{table}[bht]
    \centering

    \begin{tabular}{lccc}
        \toprule
        \textbf{Strategy}
        & \textbf{Hit@5} 
        & \textbf{R@10} 
        & \textbf{MRR@10} \\
        \midrule
        All
        & 81.11
        & 84.23
        & 70.26 \\
        
        Reverse
        & 79.64
        & 78.21
        & 67.14 \\
        
        Random
        & 82.91
        & 84.47
        & 71.36 \\
        
        \textbf{SkillSight}
        & \textbf{86.04}
        & \textbf{86.23}
        & \textbf{74.02} \\
        \bottomrule
    \end{tabular}
 \caption{Retrieval performance of different generic token selection strategies on SRA-Bench. All, Reverse, and Random estimate the background structure using all tokens, high-IDF tokens, and randomly selected tokens, respectively, while SkillSight uses low-IDF generic tokens.}
\label{tab:ordering_strategy}
\end{table}

\begin{figure}[bht]
\centering
\includegraphics[width=1\linewidth]{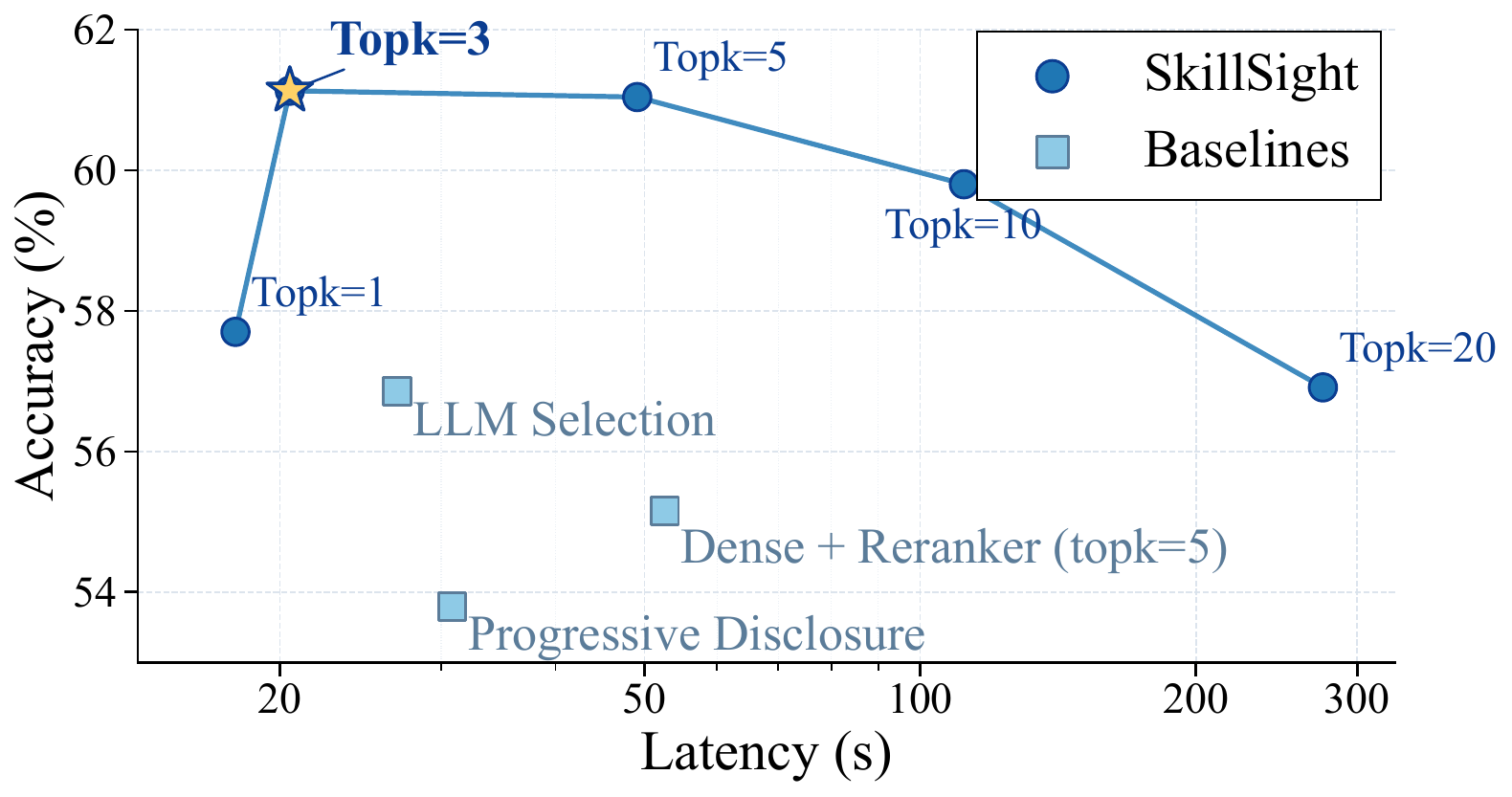}
\caption{
End-to-end accuracy--latency trade-off of different methods on SRA-Bench using Qwen3-4B-Instruct. Circles denote SkillSight with different Top-$k$ settings, while squares denote the baselines. The star marks the default Top-$3$ setting. SkillSight uses Top-$3$ in the main experiments.
}
\label{fig:4}
\end{figure}

\subsection{Ablation Study}
\label{sec:ablation}

\paragraph{Effects of the Calibration Components}
Table~\ref{tab:ablation} shows that both components provide
effective signals beyond their respective retrieval paradigms.
Adding LEC to Dense retrieval increases
$\mathrm{Recall}@10$ from 66.02 to 79.76, demonstrating that
exact matches on discriminative tokens recover fine-grained
capability differences overlooked by dense representations.
Adding SBC to BM25 yields an 11.90-point gain in
$\mathrm{Recall}@10$, indicating that calibrated semantic
signals alleviate the sensitivity of lexical retrieval to
surface-form variation.

Both components contribute to the final performance. Relative to Dense + LEC, replacing the original dense score with SBC improves Recall@10 by 6.47 points. Relative to SBC alone, adding LEC yields a further 8.00-point improvement. These results show that semantic background calibration and discriminative lexical matching provide complementary gains, with LEC contributing a larger marginal improvement under the current configuration.

\paragraph{Generic Token Selection}
As shown in Table~\ref{tab:ordering_strategy}, low-IDF generic tokens achieve the best performance across all metrics. In contrast, the high-IDF Reverse strategy reduces $\mathrm{Recall}@10$ from 86.23 to 78.21, indicating that specific tokens are less suitable for background estimation. Although the All and Random strategies retain strong recall, their lower $\mathrm{Hit}@5$ and $\mathrm{MRR}@10$ suggest that generic-token selection provides a cleaner estimate of the shared background and improves top-ranked results beyond low-rank projection alone.

\paragraph{Effect of Top-$k$}
Figure~\ref{fig:4} shows that increasing $k$ from 1 to 3 improves accuracy with little additional latency, whereas larger $k$ increases both candidate interference and inference cost. We therefore use $k=3$ in the main experiments.

\section{Conclusion}
\label{sec:conclusion}
 Our analysis reveals that these patterns form a document-side background that contributes to dense relevance scores, creates an energy gap between queries and skill documents, and obscures task-relevant signals. Motivated by this finding, we proposed SkillSight, a training-free framework that calibrates the shared background in both semantic and lexical spaces. Experiments on SRA-Bench and SkillBench-Supp demonstrate consistent improvements over strong retrieval baselines, while end-to-end evaluations confirm that these gains translate into more effective skill invocation with low inference overhead. Future work may extend background calibration to dynamically evolving skill libraries and retrieval models with jointly learned background representations.





\bibliography{aaai2026}

\end{document}